\documentclass[11pt,a4paper]{article}
\usepackage[hyperref]{emnlp2020}
\usepackage{times}
\usepackage{latexsym}

\usepackage{epsfig}
\usepackage{amsmath}
\usepackage{amssymb}
\usepackage{graphicx}
\usepackage{epstopdf}
\usepackage{wrapfig}
\usepackage{subfigure}
\usepackage{caption}
\usepackage{verbatim}
\usepackage{algorithm}
\usepackage{algpseudocode}
\usepackage{multirow}
\usepackage{booktabs}
\usepackage{amssymb}
\usepackage{tablefootnote}[2014/01/26]
\usepackage{bm}
\usepackage{dsfont}

\usepackage{microtype}

\aclfinalcopy 

\title{Improving Task-Agnostic BERT Distillation\\ with Layer Mapping Search}

\author{Xiaoqi Jiao$^{1}$\thanks{\hspace{0.5pt} This work is done when Xiaoqi Jiao is an intern at Huawei Noah's Ark Lab.}, Huating Chang$^{2}$, Yichun Yin$^{3}$, Lifeng Shang$^{3}$ \\ 
	\textbf{Xin Jiang$^{3}$, Xiao Chen$^{3}$, Linlin Li$^{4}$, Fang Wang$^{1}$ and Qun Liu$^{3}$} \\
  $^{1}$Huazhong University of Science and Technology\\
  $^{2}$Zhejiang University \\
  $^{3}$Huawei Noah’s Ark Lab \\
  $^{4}$Huawei Technologies Co., Ltd.}

\date{}

\begin{document}
\maketitle

\begin{abstract}
Knowledge distillation~(KD) which transfers the knowledge from a large teacher model to a small student model, has been widely used to compress the BERT model recently. Besides the supervision in the output in the original KD, recent works show that layer-level supervision is crucial to the performance of the student BERT model. However, previous works designed the layer mapping strategy heuristically~(e.g., uniform or last-layer), which can lead to inferior performance. In this paper, we propose to use the genetic algorithm~(GA) to search for the optimal layer mapping automatically. To accelerate the search process, we further propose a proxy setting where a small portion of the training corpus are sampled for distillation, and three representative tasks are chosen for evaluation. After obtaining the optimal layer mapping, we perform the task-agnostic BERT distillation with it on the whole corpus to build a compact student model, which can be directly fine-tuned on downstream tasks. Comprehensive experiments on the evaluation benchmarks demonstrate that 1) layer mapping strategy has a significant effect on task-agnostic BERT distillation and different layer mappings can result in quite different performances; 2) the optimal layer mapping strategy from the proposed search process consistently outperforms the other heuristic ones; 3) with the optimal layer mapping, our student model achieves state-of-the-art performance on the GLUE tasks.
\end{abstract}

\section{Introduction}
Pre-trained language models~(PLMs) such as BERT~\cite{devlin2019bert}, have achieved great success in natural language processing tasks. 
Many efforts, such as XLNet~\cite{yang2019xlnet}, RoBERTa~\cite{liu2019roberta}, ALBERT~\cite{lan2019albert}, T5~\cite{raffel2019exploring} and ELECTRA~\cite{clark2020electra}, have been paid to improve BERT with larger corpus, more parameters, or carefully designed pre-training tasks. However, these PLMs are computationally extensive and difficult to be deployed on resource-restricted devices.

Knowledge Distillation~(KD)~\cite{hinton2015distilling,romero2014fitnets} has been shown to be effective for BERT compression~\cite{tang2019distilling,sun2019patient,turc2019well,jiao2019tinybert,wang2020minilm}. 
In practice, task-agnostic distillation is desirable since it only needs to be done once and the distilled student can be applied to various downstream tasks by simply fine-tuning.
Different from original KD~\cite{hinton2015distilling} that learns the softened output of teacher network, existing task-agnostic BERT distillation methods usually utilize layer-wise supervision~\cite{jiao2019tinybert,sunmobilebert}, as knowledge embedded in the attention and hidden states of Transformer~\cite{vaswani2017attention} layers are shown to bring superior improvement on the BERT distillation. Nevertheless, there are still two critical problems unsolved:

\begin{figure}
	\centering
	\includegraphics[scale=0.31]{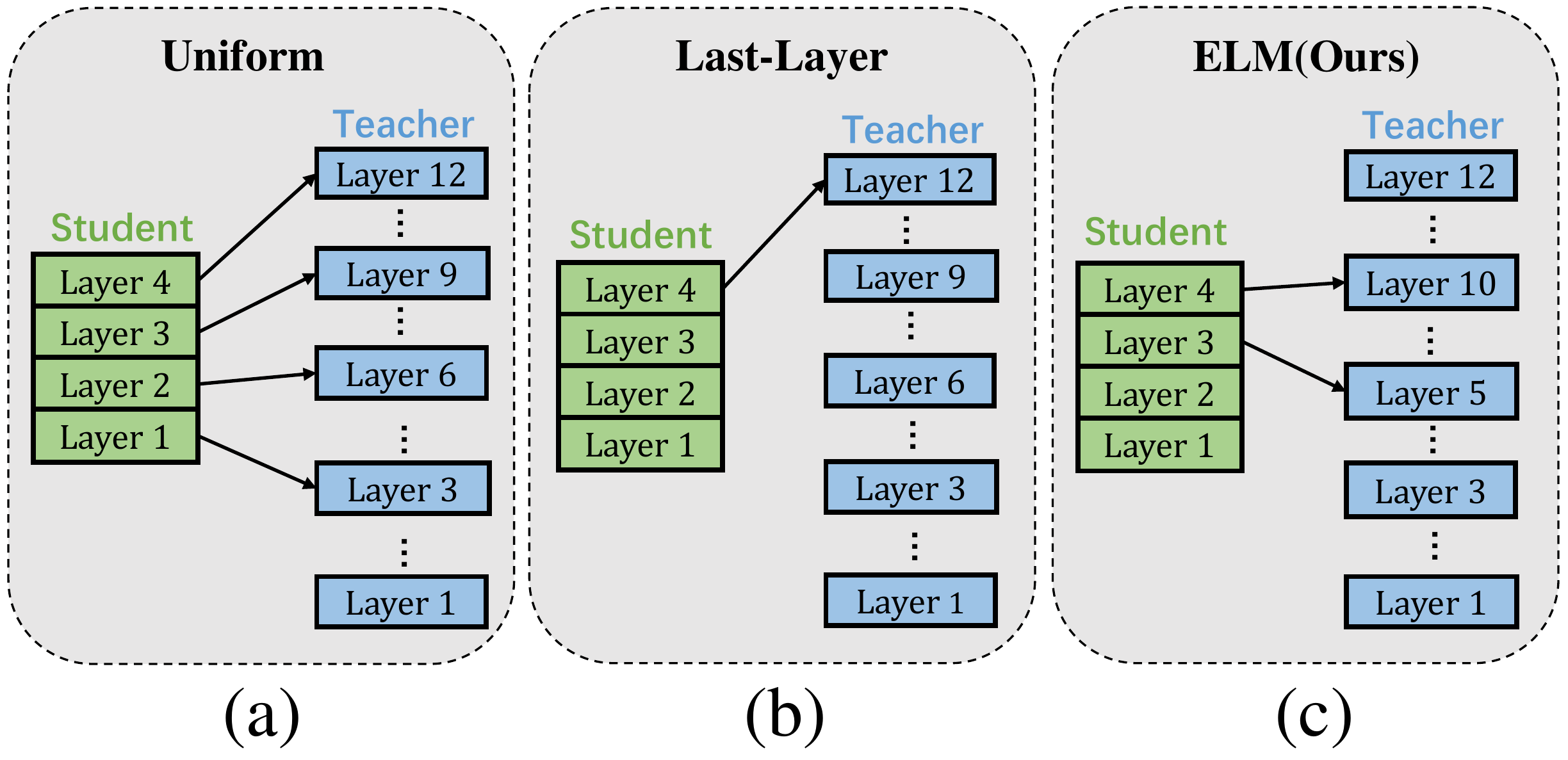}
	\caption{The diagram of different layer mapping strategies.}
	\label{figure:diagram_elm}
\end{figure}

\begin{figure*}
	\begin{center}
		\includegraphics[scale=0.5]{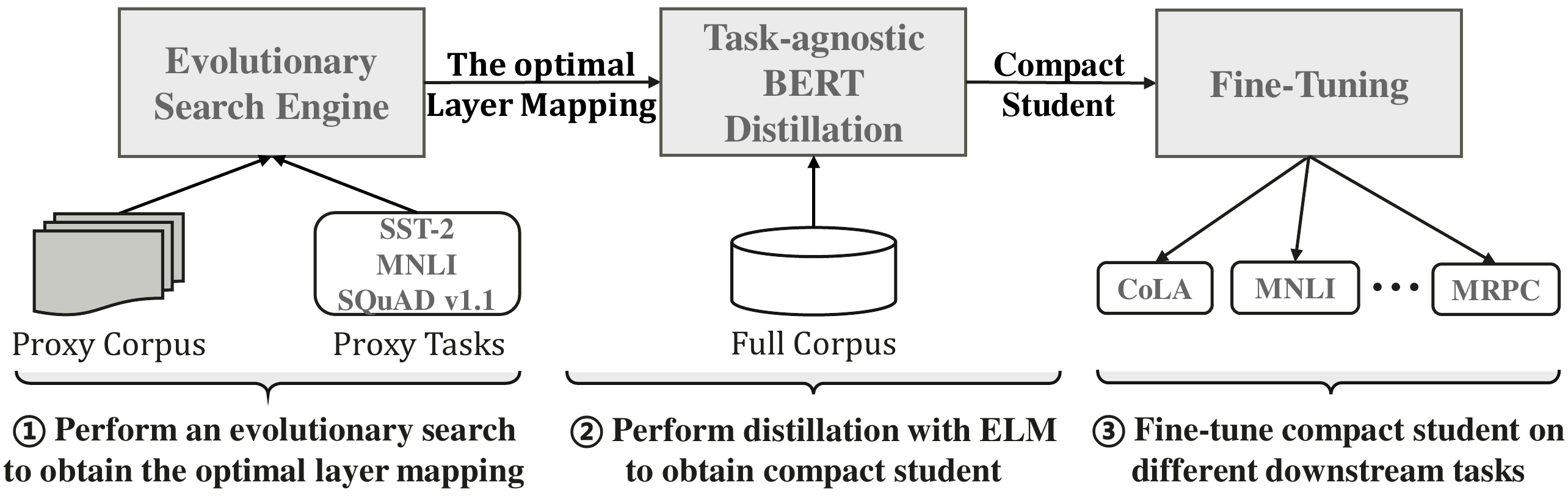}
		\caption{Overview of the layer mapping search for task-agnostic BERT distillation. We focus on the first stage to explore better layer mappings by the proposed approach under a proxy setting.}
		\label{figure:learningproceess}
	\end{center}
\end{figure*}

1) \textit{Are the current hand-crafted heuristics of layer-wise mapping optimal for BERT distillation?}
To distill from the intermediate layers of a deeper teacher model to a shallower student model, we have to select a subset of teacher's layers for supervision. Most previous works designed this important component in a hand-crafted manner. For instance, TinyBERT~\cite{jiao2019tinybert} adopts a uniform strategy that maps the teacher layers to student layers every few layers, as shown in Figure~\ref{figure:diagram_elm}(a). Another common strategy, termed as ``Last-Layer", distills knowledge from the teacher's last Transformer layer to the student's last Transformer layer~\cite{wang2020minilm}. However, these hand-crafted heuristics might still be far from being optimal. Some recent works~\cite{liu-etal-2019-linguistic,rogers2020primer} demonstrate that the intermediate layers are more transferable, which could be potentially helpful to learn generalizable knowledge in task-agnostic distillation.

2) \textit{If better layer-mapping strategies for BERT distillation exist, how can we effectively discover them?}
It is rather challenging to find the optimal layer mapping in the setting of task-agnostic BERT distillation due to the huge computational overhead. On one hand, the search space of layer mapping is large, e.g., there is a total of 9375 possibilities in distilling from a 12-layer BERT base model to a 6-layer student model. On the other hand, distillation under a certain mapping on the full pre-training corpus is also extensively time-consuming~(more than 560 V100 GPU hours for a single run in the search space). 

To tackle the aforementioned challenges, we propose the \textit{Evolved Layer Mapping}~(ELM), an automatic and efficient approach to discover the optimal layer-mapping strategy for task-agnostic BERT distillation. Specifically, ELM is armed with the genetic algorithm~(GA) and acts as an evolutionary search engine to iteratively provide layer-mapping strategies and explore better off-springs given their performance. 
To accelerate the search process, we design a proxy setting where 1) a small portion of the full corpus, called the proxy corpus, are sampled for the BERT distillation; 2) three representative tasks are chosen as the proxy tasks to evaluate layer mapping strategies. After completing the search, we obtain an optimal layer mapping, with which we perform the task-agnostic BERT distillation on the whole corpus to get a compact student, and compare it with student models, distilled with other widely-used mapping strategies via directly fine-tuning on downstream tasks. The whole workflow is shown in Figure~\ref{figure:learningproceess}.

In summary, the contributions of this work are three-fold: 
\begin{itemize}
\item To the best of our knowledge, we are the first to systematically study the effect of layer mapping strategies in BERT Distillation.
\item We propose ELM, an automatic and efficient approach to search the optimal layer mapping strategy.
\item Extensive experiments on benchmarks demonstrate that our ELM obtains consistently better results than the widely-used heuristics such as the uniform or last-layer distillation, and outperform the state-of-the-art baselines.
\end{itemize}

\section{Related Work}
Generally, pre-trained language models~(PLMs) compression can be categorized into low-rank approximation~\cite{ma2019tensorized,Lan2020ALBERT}, weight sharing~\cite{dehghani2018universal,Lan2020ALBERT}, knowledge distillation~\cite{tang2019distilling,sanh2019distilbert,turc2019well,sun2020mobilebert,liu2020fastbert,wang2020minilm}, pruning~\cite{cui2019fine,mccarley2019pruning,Fan2020Reducing,Elbayad2020Depth-Adaptive,gordon2020compressing,hou2020dynabert,zhang2020know} and quantization~\cite{shen2019q,zafrir2019q8bert}.

Knowledge distillation~(KD), an idea proposed by Hinton et al. \cite{hinton2015distilling} in a teacher-student framework, has been proven to be a promising approach to compress large pre-trained language models~(PLMs). Typically, KD methods for PLMs can be summarized into task-agnostic distillation at the pre-training stage and task-specific distillation at the fine-tuning stage. Task-specific distillation~\cite{tang2019distilling,sun2019patient,turc2019well} uses a fine-tuned BERT as the teacher and performs distillation on the task dataset to make student learn the knowledge for a specific task. Task-agnostic distillation~\cite{tsai2019small,sanh2019distilbert,sunmobilebert,wang2020minilm}, on the contrary, uses pre-trained BERT as the teacher and performs distillation on large unsupervised corpus to make the student acquire general knowledge. The distilled student model is task-agnostic which can be easily applied to different downstream tasks by fine-tuning. In addition, TinyBERT~\cite{jiao2019tinybert} combines these two kinds of methods and shows that it can further improve the performance of student models. 

Existing BERT distillation methods typically focus on: a) defining new types of knowledge such as attention, hidden states~\cite{jiao2019tinybert,sunmobilebert}, and value-relation~\cite{wang2020minilm}; and b) introducing new processes of knowledge transfer, such as joint knowledge transfer, progressive knowledge transfer~\cite{sunmobilebert} and progressive module replacement~\cite{xu2020bert}. In this work, we focus on improving the task-agnostic distillation with a better layer mapping obtained by the proposed efficient search method.

The genetic algorithm~(GA)~\cite{mitchell1998introduction,sadeghi2014optimizing} is a meta-heuristic method inspired by the natural selection process. It is commonly used to generate high-quality solutions to optimization and search problems by performing bio-inspired operators such as mutation, crossover, and selection. It is simple and often applied as an approach to solve global optimization problems. Generally, the GA has two components: a genetic representation of the search space and a fitness function that is used to evaluate the representations. Previous works applied the genetic algorithm to 1) prune a large neural networks\cite{BEBIS1997167,WANG2020247}; 2) automatically explore the neural network architectures both in the computer vision~\cite{yao1999evolving,xie2017genetic} and natural language processing~\cite{so2019evolved,liu2020evolving}; 3) conduct feature selection in machine learning\cite{ZENG20091214,KABIR20112914}.
Besides, researchers also proposed to extract linguistic rules from data sets using fuzzy logic and genetic algorithms~\cite{MENG201248}. Concurrently, BERT-EMD~\cite{li2020bert} investigates Many-to-Many Layer Mapping for BERT Compression on downstream tasks with Earth Mover's Distance. In this paper, we try to find the optimal mapping function of task-agnostic BERT distillation with the help of the genetic algorithm.

\section{Preliminaries}
In this section, we will briefly describe the formulation of Transformer~\cite{vaswani2017attention} Encoder layer, Knowledge Distillation~\cite{hinton2015distilling} and its application on Transformer-based models. Then, we illustrate the problem that we are going to investigate in this paper.

\subsection{Transformer Encoder Layer}
Most recently proposed pre-trained language models (e.g., BERT and RoBERTa) are built on Transformer encoder layers, which has been proved to capture long-term dependencies between the input tokens by the self-attention mechanism. Specifically, a Transformer encoder layer consists of two sub-layers: {\it multi-head attention}~(MHA) and {\it fully connected feed-forward} network~(FFN).

\noindent \textbf{Multi-Head Attention (MHA)}. The attention mechanism depends on the three components: queries ($\bm{Q}$), keys ($\bm{K}$) and values ($\bm{V}$), and can be formulated as follows:
\begin{align}
	\label{eq:attention_score}
	\!\!\bm{A} \! & = \! \frac{\bm{Q}\bm{K}^{T}}{\sqrt{d_k}}, \\
	\!\!\texttt{Attention}(\bm{Q},\bm{K},\bm{V}) \! & = \! \texttt{softmax}(\bm{A})\bm{V},
\end{align}
where $d_k$ is the dimension of keys, $\bm{A}$ is the attention matrix calculated from the compatibility of $\bm{Q}$ and $\bm{K}$ by dot-product operation. The final function output is calculated as a weighted sum of values $\bm{V}$ where the weight can be obtained by applying {\tt softmax()} on $\bm{A}$.

 
Multi-head attention concatenates the attention heads from different representation subspaces and introduces a linear transformation $\bm{W}$ to combine the information from these subspaces. It is defined as follows:
\begin{align}
\label{muti-head-attention}
\texttt{MHA}(\bm{Q},\bm{K},\bm{V}) \! &= \! \texttt{Concat}({\rm h}_1,\ldots,{\rm h}_k)\bm{W},
\end{align}
where $k$ is the number of attention heads, and ${\rm h}_i$ denotes the $i$-th attention head, which is calculated by the $\texttt{Attention}()$ function with inputs from different representation subspaces. 

\noindent \textbf{Position-wise Feed-Forward Network (FFN)}. In addition to the multi-head attention sub-layer, each transformer encoder layer also contains a fully connected feed-forward network, which is formulated as follows:
\begin{align}
\label{eq:FFN}
\texttt{FFN}(x) = \max(0, x\bm{W}_1 + b_1)\bm{W}_2 +b_2.
\end{align}
The FFN contains two linear transformations and one ReLU activation and it is applied to each token separately and identically.

\subsection{Knowledge Distillation}
Knowledge Distillation~\cite{hinton2015distilling} (KD) aims to transfer the knowledge of a large teacher network $T$ to a small student network $S$. The knowledge transfer can be done by making student model mimic the behaviors of teacher model, which is formulated as minimizing the following objective:
\begin{align}
\label{eq:point_distillation}
\mathcal{L}_{KD} = \sum_{x \in \mathcal{X}} L\big(S(x), T(x)\big),
\end{align}
where $L$ is a loss function, $\mathcal{X}$ denotes the dataset, $T(x)$ and $S(x)$ mean the behavior functions of teacher and student networks on data sample $x$ respectively. Original KD defines the softened output as the behavior.

\subsection{Task-agnostic BERT Distillation}
Task-agnostic BERT distillation aims to learn a compact student model, which can be directly fine-tuned on downstream tasks as the pre-trained BERT teacher does. Different from the original KD, it usually adopts the transformer-layer supervision to guide the student model, which is proved to be effective~\cite{jiao2019tinybert,sunmobilebert}. Assuming the teacher model has $N$ Transformer layers and the student model has $M$ Transformer layers, we need to design a mapping function $g(m)$ to choose some specific layers of the teacher model for the layer-wise learning. By introducing this layer-wise supervision into the BERT distillation, the new objective is formulated as:
\begin{align}
\label{eq:kd_loss}
\mathcal{L}_{KD} = \sum_{x \in \mathcal{X}} \sum\nolimits^{M}_{m=1} \!\lambda_{m} L_{\text{layer}}(S_m(x), T_{g(m)}(x)),
\end{align}
where $S_m$ and $T_{g(m)}$ refer to the $m$-th layer of student and the $g(m)$-th layer of teacher, the parameter $\lambda_m \in \left\lbrace 0, 1 \right\rbrace $ denotes whether the $m$-th layer of student learns from a layer of teacher or not. Task-agnostic BERT distillation uses the large-scale unsupervised corpus as the dataset $\mathcal{X}$. There are several effective behavior functions in the previous works~\cite{jiao2019tinybert,wang2020minilm}. We adopt the behavior function from TinyBERT~\cite{jiao2019tinybert} and therefore the loss for $m$-th Transformer layer of the student $L_{\text{layer}}(S_m, T_{g(m)})$ can be formulated:
\begin{align}
\label{eq:transformer_layer_loss}
\frac{1}{h}\sum^{h}_{i=1} \left\|\bm{A}_i^{S_m} \!\!-\!\! \bm{A}_i^{T_{g(m)}}\right\|_2^2 \!+\! \left\|\bm{H}^{S_m}\bm{W}_h \!\!-\!\! \bm{H}^{T_{g(m)}}\right\|_2^2,
\end{align}
where $h$ is the number of attention heads, $\bm{A}_i^{S_m}$ and $\bm{A}_i^{T_{g(m)}}$ refer to the attention matrix corresponding to the $i$-th head of $m$-th layer of student and $g(m)$-th layer of teacher network respectively. Similarly, $\bm{H}^{S_m}$ and $\bm{H}^{T_{g(m)}}$ refer to the hidden states of $m$-th layer of student and $g(m)$-th layer of teacher network respectively. The learnable matrix $\bm{W}_h$ transforms the hidden states of student model into the same space of teacher model. 

\subsection{Layer Mapping Search} 
There are many possible layer mappings for task-agnostic BERT distillation. Previous works~\cite{jiao2019tinybert,wang2020minilm} design the layer mapping function heuristically, such as uniform and top-layer as illustrated in Figure~\ref{figure:diagram_elm}, which inevitably introduces human design bias and leads to inferior performance. Layer mapping search aims to find the optimal $g(m)$ in Equation \ref{eq:kd_loss} from all the possible ones.

\begin{figure*}
	\centering
	\includegraphics[scale=0.48]{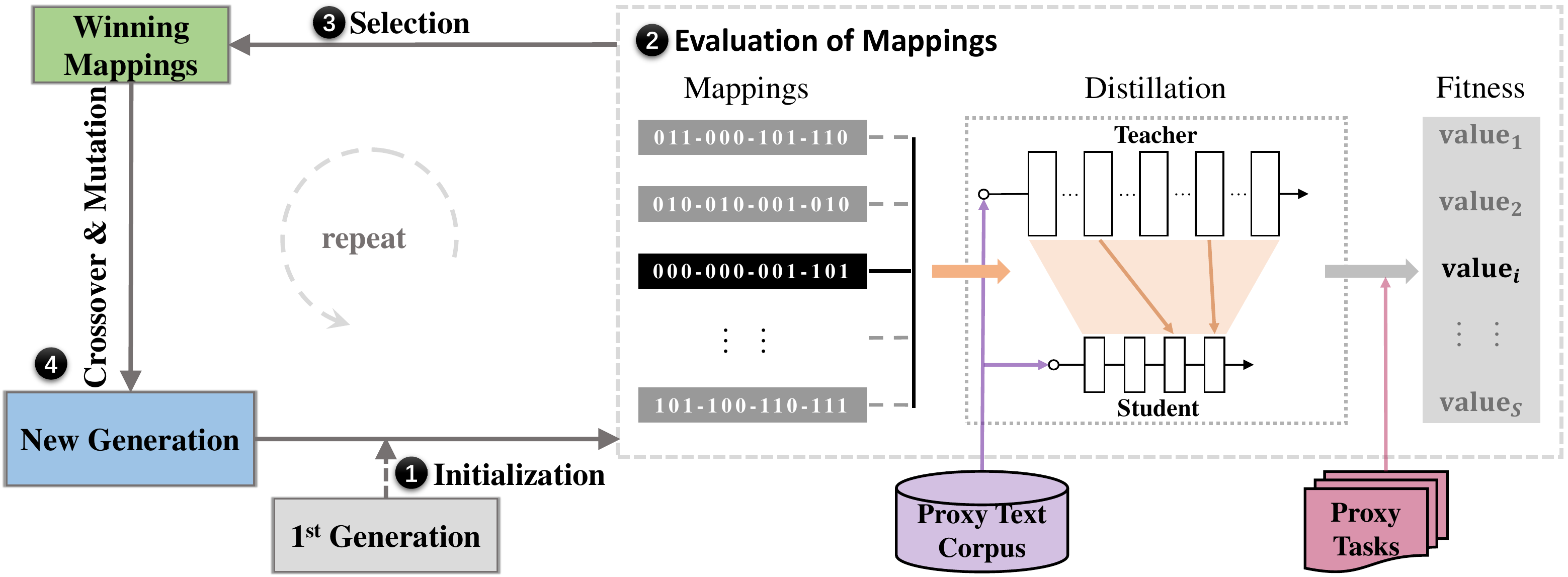}
	\caption{The overview of ELM~(or the evolutionary search engine). The process includes four stages: 1) Start with a randomly initialized \textit{generation}, the first generation, which consists of a set of \textit{genes}. 2) Run the task-agnostic BERT distillation in parallel with different \textit{genes}~(layer mappings) on the proxy corpus to obtain corresponding students. 3) Pair each \textit{gene} with a \textit{fitness} value by fine-tuning the corresponding student on some representative proxy tasks. 4) Perform the genetic algorithm~(GA) to select the \textit{genes} and produce a new generation by the genetic operations \textit{crossover} and \textit{mutation}. By repeating stage 2 to 4, ELM would evolve better layer mappings automatically and find the optimal \textit{gene}.}
	\label{figure:overview}
\end{figure*}

\section{Method}
\subsection{Overview}
An overview of the ELM is shown in Figure~\ref{figure:overview}. We firstly generate a set of randomly initialized \textit{genes}~(layer mappings). Then, we start the evolutionary search engine: 1) Perform the task-agnostic BERT distillation with \textit{genes} in the current generation to obtain corresponding students. 2) Get the fitness value by fine-tuning each student on the proxy tasks. 3) Apply the genetic algorithm to evolve the next generation.
 
\subsection{Search Space and Gene Encoder} In the GA framework, we treat each layer mapping as a gene that is formatted as a binary string. The mapping function $n = g(m)$ is used to determine where the $m$-th layer of the student model should learn information from the teacher model. Then a layer mapping can be denoted as a tuple $(g(1),g(2),\cdots,g(M))$. We design a search space for each element in the layer mapping, which is formulated as:
\begin{equation}
\!g(m) \! \in \!
\begin{cases}
\left[Z\!+\!1, Z\!+\!2^{k}\!-\!1 \right] \! \cup \! \{None\}  & m \!\!< \!\!M \\
\left[N\!-\!2^{k}+1, N \right]            & m\!\! =\!\! M \end{cases},
\label{eq:model_loss}
\end{equation}
where $Z = \left \lfloor  \frac{(m-1) \cdot N}{2 \cdot M} \right \rfloor$ and each layer in the student model has a search space of $2^{k}$.
For non-last elements, we allow them not to learn any layer of teacher model, which is denoted as {\it None}. To adapt the gene encoding for layer mappings, we choose $k$ as the minimum value that subjects to $2^{k}>\frac{2 \cdot N}{M}$. We give some examples of whole search spaces for different teacher-student settings in Table~\ref{tab:search space}.
\begin{table}[h]
	\caption{Search spaces for different (student, teacher) architectures. Note that the spaces for non-last layers contain {\it None}, we do not add it to the table for clarity.}
	\label{tab:search space}
	\vspace{-0.15in}
	\begin{center}
		\scalebox{0.75}{
			\begin{tabular}{cc|cccccc|c}
				\hline
				M & N  &  g(1)  &  g(2)  &  g(3)  &  g(4)  &  g(5)  &  g(6)  & Space  \\ \hline
				4 & 12 & [1,7]  & [2,8]  & [4,10] & [5,12] &   -    &   -    &  1048  \\
				6 & 12 & [1,7]  & [2,8]  & [3,9]  & [4,10] & [5,11] & [5,12] &  9375  \\
				4 & 24 & [1,15] & [4,18] & [7,21] & [9,24] &   -    &   -    & 13892  \\
				6 & 24 & [1,15] & [3,17] & [5,19] & [7,21] & [9,23] & [9,24] & 380321 \\ \hline
		\end{tabular}}
	\end{center}
\vspace{-0.1in}
\end{table}

\begin{algorithm*}[ht]
	\caption{The Genetic Algorithm for Layer Mapping Search}
	\begin{algorithmic}[1]
		\State \textbf{Input}: the number of generations $T$, the number of \textit{genes} in each generation $S$, the
		mutation and crossover probabilities $p_{M}$ and $p_{C}$, the mutation and crossover parameter $q_{M}$ and $q_{C}$, The sampled training corpus $\mathbb{C}$ and evaluation datasets $\mathbb{D}$.
		\State Initialize $\mathbb{G}_{1}$, perform task-agnostic BERT distillation with the \textit{genes} in it on $\mathbb{C}$, obtain $\mathbb{V}_{1}$ by evaluating them on $\mathbb{D}$.
		\For{$t=2, 3 \cdots, T$}
		\State $\mathbb{G}_{t} \gets  \{\} $
		\While{$|\mathbb{G}_{t}|<S$}
		\State Sample two genes: ${\rm g_1}$, ${\rm g_2}$ with a Russian roulette process on $\mathbb{G}_{t-1}$.
		\State With probability $p_{C}$, do crossover for pair (${\rm g_1}$, ${\rm g_2}$) under the parameter $q_{C}$.
		\State With probability $p_{M}$, do mutation for ${\rm g_1}$ and ${\rm g_2}$ individually under the parameter $q_{M}$.
		\State Append the newly generated or original sampled genes into $\mathbb{G}_{t}$.
		\EndWhile
		\State Obtain $\mathbb{V}_{t}$  for $\mathbb{G}_{t}$ by perform the distillation and evaluation process.
		\EndFor
		\State \textbf{Output}: Output the gene with best fitness in $\mathbb{G}_{T}$ as the Evolved Layer Mapping.
	\end{algorithmic}
	\label{alg:ga}
\end{algorithm*}

With $k$-bit to represent a mapping element, taking $k=3$ for example, the first layer in the search space is represented by a binary string `-001-' and the second one is represented by '-010-', etc. In addition, we represent {\it None} as '-000-'. By concatenating the binary strings of all elements in the tuple, we get a binary representation for a specific layer mapping. We do not allow the cross in the search space, namely for $1\leq m_i<m_j\leq M$, we ensure $1\leq g(m_i)<g(m_j)\leq N$.

\subsection{Search under A Proxy Setting}
Since performing task-agnostic BERT distillation with a certain mapping on the full pre-training corpus is time-consuming~(more than 560 V100 GPU hours for student models with 6 layers), we design a proxy setting, under which 10\% of the whole corpus are sampled for BERT distillation during the search process. By putting the GA under the proxy setting, we can reduce the computational expense by about 10x. However, a potential concern is whether proxy training is sufficient to represent the standard training with full corpus, as there is a performance gap between them. We shall empirically investigate this by observing {\it the relative performance order} property. Later, in the experimental section, we will verify the effectiveness of the proxy setting. 

Moreover, to evaluate the layer mappings during the search process, we fine-tune their corresponding students on three representative proxy tasks: sentiment analysis, natural language inference, and extractive question answering.

\subsection{Efficient Layer Mapping Search via GA}
Enumerating all possible layer mappings under the task-agnostic BERT distillation setting is intractable because of 1) the large space of layer mappings, as shown in Table~\ref{tab:search space}; 2) the large scale pre-training corpus used for distillation. To alleviate the expensive computational overhead, we introduce the genetic algorithm to efficiently explore the search space under the aforementioned proxy setting. In the following, we detail the genetic algorithm for layer mapping search.


{\it Initialization.} We initialize a set of valid randomized layer mapping $\{\mathbb{G}_{1}^s\}_{s=1}^{S}$ as the first generation, where each layer mapping is formated as a \textit{gene} and each bit in it is independently sampled from a Bernoulli distribution $\mathbb{B}(0.5)$. 

{\it Evaluation and Selection.} After performing the task-agnostic BERT distillation with all the \textit{genes} within a generation, we assign each of them $\mathbb{G}_{t}^s$ a fitness value $\mathbb{V}_{t}^s$, which is the average performance by fine-tuning the corresponding student on the proxy evaluation tasks. Then we perform a Russian roulette process\footnote{\url{https://en.wikipedia.org/wiki/Fitness_proportionate_selection}} to determine which genes survive. The survival probability of gene $\mathbb{G}_{t}^s$ is proportional to $\mathbb{V}_{t}^s - \min{\{\mathbb{V}_{t}^s\}}_{s=1}^{S}$, where $\min{\{\mathbb{V}_{t}^s\}}_{s=1}^{S}$ is the minimal fitness value in the generation.

{\it Mutation and Crossover.} We conduct mutation of a gene by flipping each bit independently with a probability $q_{M}$. Practically, $q_{M}$ is often small so that a gene would not be changed too much and its good properties can be preserved. In the crossover process, we exchange the corresponding mapping elements in two \textit{genes} with a probability $q_{C}$. By crossover, we can produce new genes and keep the good mapping elements in the next generation. 

By conducting these genetic operations, we can obtain the next generation. The probabilities that we perform mutation and crossover are $p_{M}$ and $p_{C}$, respectively. The genetic algorithm for layer mapping search is illustrated in Algorithm~\ref{alg:ga}.

\section{Experiments}
In this section, we empirically study the effects of the proposed ELM approach on task-agnostic BERT distillation. We first compare the layer mapping strategies found by our ELM to other handcrafted heuristics and analyze these strategies in detail. Then we compare ELM to various state-of-the-art results. Finally, we also provide a set of further analysis to shed more light on ELM.


\begin{table*}[]
	\caption{Comparison between student models with different layer mappings on the dev set. The fine-tuning results are averaged over 3 runs. The 4-layer student has an architecture of ($M$=4, $d$=312, $d_i$=1200, $h$=12), and the 6-layer student has an architecture of ($M$=6, $d$=384, $d_i$=1536, $h$=12).} \label{tab:main_results}
	\centering
	\scalebox{0.78}{
		\begin{tabular}{l|l|c|ccc|ccccc|c}
			\toprule
			\multirow{2}{*}{Student} & \multirow{2}{*}{Strategy} & \multirow{2}{*}{Layer Mapping} &     SST-2     &     MNLI      &     SQuAD v1.1     &        MRPC        &     CoLA      &     QNLI      &        QQP         &     SQuAD v2.0     & \multirow{2}{*}{Avg} \\
			&                           &                                &     (acc)     &     (acc)     &      (EM/F1)       &      (acc/F1)      &     (mcc)     &     (acc)     &      (acc/F1)      &      (EM/F1)       &                      \\ \midrule
			\multirow{4}{*}{4-Layer} & Uniform                   &           (3,6,9,12)           &     87.4      &     77.0      &     66.7/77.4      &     76.5/84.6      &     21.3      &     84.9      &     86.0/81.7      &     58.9/62.5      &         72.1         \\
			& Last-Layer                &           (0,0,0,12)           &     88.1      &     77.6      &     69.2/79.4      &     83.8/88.6      &     21.4      &     85.7      &     87.2/82.9      &     59.4/63.3      &         73.4         \\
			& Contribution              &          (1,10,11,12)          &     86.8      &     76.1      &     64.4/76.4      &     79.4/86.3      &     15.5      &     85.8      &     86.1/81.4      &     61.6/65.1      &         71.7         \\
			& ELM (ours)                &           (0,0,5,10)           & \textbf{89.9} & \textbf{78.6} & \textbf{71.5/81.2} & \textbf{85.0/89.5} & \textbf{23.9} & \textbf{86.0} & \textbf{87.9/83.6} & \textbf{62.9/66.2} &    \textbf{74.9}     \\ \midrule
			\multirow{4}{*}{6-Layer} & Uniform                   &        (2,4,6,8,10,12)         &     90.7      &     81.2      &     76.0/84.6      &     85.0/89.6      &     27.2      &     89.2      &     88.2/84.1      &     68.0/71.3      &         77.2         \\
			& Last-Layer                &         (0,0,0,0,0,12)         &     89.8      &     81.3      &     76.0/84.7      &     85.5/89.7      &     34.3      &     89.0      &     88.7/84.6      &     68.7/71.9      &         78.2         \\
			& Contribution              &        (1,6,7,10,11,12)        &     90.0      &     80.9      &     75.0/84.1      &     84.6/89.3      &     28.5      &     88.8      &     88.0/84.2      &     66.5/70.0      &         77.0         \\
			& ELM  (ours)               &         (0,5,0,0,0,10)         & \textbf{91.5} & \textbf{82.4} & \textbf{77.2/85.7} & \textbf{86.0/90.1} & \textbf{36.1} & \textbf{89.3} & \textbf{89.2/85.4} & \textbf{70.3/73.2} &    \textbf{79.2}     \\ \bottomrule
	\end{tabular}}
\end{table*}

\subsection{Datasets}
\label{subsec:english_exp}
For task-agnostic BERT distillation, we take the BooksCorpus and English Wikipedia as the unsupervised text corpus. 
The preprocessing follow the standard procedures in BERT tokenization~\cite{devlin2019bert}.
After task-agnostic distillation, we fine-tune the student model on the GLUE benchmark~\cite{wang2018glue}, including two single-sentence tasks: sentiment analysis~(SST-2) and linguistic acceptability~(CoLA), two natural language inference tasks~(MNLI, QNLI) and two sentence similarity tasks~(MRPC and QQP). Besides, we also evaluate the ELM on two extractive question answering tasks: SQuAD v1.1 and v2.0. 


\subsection{Implementation Details}

\paragraph{Hyperparameter Setting}
For the teacher model, we use the pre-trained BERT$_{\rm BASE}$ with 12 layers ($N$=12, $d$=768, $d_i$=3072, $h$=12)\footnote{$N$ (or $M$) refer to the layer numbers of teacher (or student) models, $d$ and $d_i$ mean the hidden and feed-forward/filter size respectively, $h$ means the head number.}. To instantiate student models, we adopt two small BERT architectures: 1) a 4-layer model with $M$=4, $d$=312, $d_i$=1200, $h$=12, and 2) a 6-layer model with $M$=6, $d$=384, $d_i$=1536, $h$=12. Our implementation is built upon TinyBERT\footnote{https://github.com/huawei-noah/Pretrained-Language-Model/tree/master/TinyBERT}, but equipped with different mapping functions found by ELM. 
In the stage of task-agnostic distillation, we set the batch size to 256, with a peak learning rate of 1e-4. We set the maximum sequence length of 128 for the first three epochs, and increase it to 512 for two more epochs. The other hyper-parameters are the same as the original BERT~\cite{devlin2019bert}. In the fine-tuning stage on GLUE tasks, we set the batch size to 32, and enumerate the learning rate from \{1e-5, 2e-5, 3e-5\} and epoch from \{3, 5, 10\}. The maximum sequence length is set to 64 for single sentence tasks, and 128 for other sequence pair tasks. For question answering tasks~(SQuAD v1.1 and v2.0), the maximum sequence length is 384, and we fine-tune the distilled models for 4 epochs, using 16 as the batch size and 3e-5 as the learning rate.

\paragraph{Setup of Genetic Algorithm}
To reduce the search cost, we perform the GA under a proxy setting where 10\% of the whole corpus are sampled for the task-agnostic BERT distillation. This can bring approximately $10\times$ time reduction in the distillation stage. Later, we shall empirically investigate the effectiveness of this proxy setting.
We run the genetic search for 5 generations, and the number of genes in each generation is 12 and 20 respectively for the 4-layer and 6-layer student network. 
Unless otherwise specified, the parameters of genetic algorithm are set to $p_{M}$ = 0.8, $q_{M}$ = 0.05, $p_{C}$ = 0.2 and $q_{C}$ = 0.2 in this paper. 

To get the fitness scores of genes during the search process, we take the average performance of three representative tasks (SST-2, MNLI, and SQuAD v1.1) for evaluation. We use F1 as the metric for SQuAD v1.1 and accuracy for the other two datasets.
Besides, we evaluate the distilled student models by fine-tuning them on the evaluation datasets with the same hyper-parameters.

\subsection{Results and Analysis}

\begin{table*}[]
	\caption{Comparisons with SOTA task-agnostic distillation baselines with architecture of (M=6, d=768, di=3072, h=12) or (M=4, d=768, di=3072, h=12) on the GLUE dev sets. The results of our fine-tuning experiments are averaged over 3 runs for each task. $^*$ means the results were not reported in the original paper, and we fine-tune the released model with the suggested hyper-parameters .$\dagger$~denotes that the results are taken from \cite{wang2020minilm}.}
	\label{tab:sota_compare_dev}
	\centering
	\scalebox{0.76}{
		\begin{tabular}{l|ccc|cccccccc|c}
			\toprule
			\multirow{2}{*}{Method}                      & \multirow{2}{*}{\#Layer} & \multirow{2}{*}{\#Param} & \multirow{2}{*}{Speedup} &     CoLA      &    MNLI-m     &     MRPC      &     QNLI      &      QQP      &      RTE      &     SST-2     &     STS-B     & \multirow{2}{*}{Avg} \\
			&                          &                          &                          &     (mcc)     &     (acc)     &     (acc)     &     (acc)     &     (acc)     &     (acc)     &     (acc)     &    (spear)    &                      \\ \midrule
			BERT$_{\rm BASE}$ (Teacher)                  &            12            &           109M           &       1.0$\times$        &     58.1      &     84.8      &     87.7      &     92.0      &     90.9      &     71.1      &     92.9      &     89.6      &         83.4         \\ \midrule
			Fine-tuning~\cite{tsai2019small}             &            6             &          67.0M           &       2.0$\times$        &   50.1$^*$    &     81.8      &     81.8      &     87.9      &     88.5      &     64.2      &     92.5      &   87.9$^*$    &         79.3         \\
			DistilBERT~\cite{sanh2019distilbert}         &            6             &          67.0M           &       2.0$\times$        &     51.3      &     82.2      &     87.5      &     89.2      &     88.5      &     59.9      &     91.3      &     86.9      &         79.6         \\
			Ta-TinyBERT$\dagger$~\cite{jiao2019tinybert} &            6             &          67.0M           &       2.0$\times$        &     42.8      &     83.5      &     88.4      &     90.5      &     90.6      & \textbf{72.2} &     91.6      &   88.5$^*$    &         81.0             \\
			MiniLM$\dagger$~\cite{wang2020minilm}        &            6             &          67.0M           &       2.0$\times$        &     49.2      &     84.0      &     88.4      & \textbf{91.0} &     91.0      &     71.5      &     92.0      &       -       &                      \\
			\textbf{ELM~(ours)}                          &            6             &          67.0M           &       2.0$\times$        & \textbf{54.2} & \textbf{84.2} & \textbf{89.0} &     90.8      & \textbf{91.1} & \textbf{72.2} & \textbf{92.7} & \textbf{88.9} &    \textbf{82.9}     \\ \midrule
			Fine-tuning~\cite{tsai2019small}             &            4             &          52.2M           &       3.0$\times$        &     33.9      &     78.4      &     86.0      &     82.3      &     87.1      &     58.2      &     87.2      &     78.4      &         73.9         \\
			DistilBERT~\cite{sanh2019distilbert}         &            4             &          52.2M           &       3.0$\times$        &     40.0      &     78.4      &     83.1      &     84.5      &     88.8      &     56.3      &     90.0      &     83.8      &         75.6         \\
			\textbf{ELM~(ours)}                          &            4             &          52.2M           &       3.0$\times$        & \textbf{42.4} & \textbf{81.6} & \textbf{87.5} & \textbf{88.5} & \textbf{90.1} & \textbf{67.5} & \textbf{91.5} & \textbf{87.8} &    \textbf{79.6}     \\ \bottomrule
	\end{tabular}}
\end{table*}

\begin{table*}[]
	\caption{Comparisons with SOTA task-agnostic distillation methods with the same architecture of (M=4, d=768, di=3072, h=12) on the GLUE test sets. $\ddagger$~denotes that the results are taken from \cite{jiao2019tinybert}.}
	\label{tab:sota_compare_test}
	\centering
	\scalebox{0.73}{
		\begin{tabular}{l|ccc|cccccccc|c}
			\toprule
			\multirow{2}{*}{Method}                        & \multirow{2}{*}{\#Layer} & \multirow{2}{*}{\#Param} & \multirow{2}{*}{Speedup} &     CoLA      &    MNLI-(m/mm)     &     MRPC      &     QNLI      &      QQP      &      RTE      &     SST-2     &     STS-B     & \multirow{2}{*}{Avg} \\
			&                          &                          &                          &     (mcc)     &       (acc)        &     (acc)     &     (acc)     &     (acc)     &     (acc)     &     (acc)     &    (spear)    &                      \\ \midrule
			BERT$_{\rm BASE}$ (Teacher)                    &            12            &           109M           &       1.0$\times$        &     52.8      &     83.9/83.4      &     87.5      &     90.9      &     71.1      &     67.0      &     93.4      &     85.2      &         79.5         \\ \midrule
			Fine-tuning~\cite{tsai2019small}               &            4             &          52.2M           &       3.0$\times$        &     31.5      &     78.3/77.4      &     82.7      &     86.1      &     68.8      &     62.2      &     90.7      &     78.8      &         72.9         \\
			DistilBERT$\ddagger$~\cite{sanh2019distilbert} &            4             &          52.2M           &       3.0$\times$        &     32.8      &     78.9/78.0      &     82.4      &     85.2      &     68.5      &     54.1      &     91.4      &     76.1      &         71.9         \\
			\textbf{ELM~(ours)}                            &            4             &          52.2M           &       3.0$\times$        & \textbf{34.9} & \textbf{80.6/79.5} & \textbf{85.9} & \textbf{86.7} & \textbf{68.9} & \textbf{63.9} & \textbf{91.9} & \textbf{80.5} &    \textbf{74.8}     \\ \bottomrule
	\end{tabular}}
\end{table*}

\subsubsection{Comparisons with Layer Mapping Heuristics}
We first verify the effectiveness of layer mapping strategies on task agnostic distillation. We compare ELM with three heuristic layer mapping strategies as follows:
\begin{itemize}
\item {\bf Uniform~\cite{jiao2019tinybert}} The student network learns evenly from the layers of the teacher network, and the corresponding layer mappings for 4- and 6-layer students are (3, 6, 9, 12) and $(2, 4, 6, 8, 10, 12)$ respectively. 
\item {\bf Last-layer~\cite{wang2020minilm}} Only the last layer of student network learns from the last layer of teacher network, and the corresponding layer mappings for 4- and 6-layer student are $(0, 0, 0, 12)$ and $(0, 0, 0, 0, 0, 12)$, respectively. 
\item{\bf Contribution~\cite{sajjad2020poor}} The contribution of each layer is defined as the cosine similarity between its input and output, and the lower the cosine value the more important the layer. We use 10\% of the training corpus to obtain the contribution scores, and obtain the mappings as $(1, 10, 11, 12)$ and $(1, 6, 7, 10, 11, 12)$ for the 4- and 6-layer students, respectively.
\end{itemize}
The comparison results are shown in Table~\ref{tab:main_results}. It can be observed that our ELMs consistently outperform the three widely-used layer mapping heuristics on all datasets. For instance, our ELM outperforms on average the contribution approach by 3.2\% and 2.2\% for the 4-layer and 6-layer settings, respectively.
It is worth mentioning that the ELM searched by the datasets of SST-2, MNLI, and SQuAD v1.1, can also be generalized on the other datasets and achieves better results than baselines.
In summary, the significant improvement demonstrates the superiority and generalization ability of our ELM search framework\footnote{We also demonstrate the effectiveness of ELM on Chinese tasks. Due to space limitation, we present the details in Appendix.}.

\paragraph{Looking into Mapping Strategies of ELM}
In Table~\ref{tab:main_results}, we take a further look at the mapping strategies found by our ELM. We have the following observations:
\textbf{1) Not every layer of student needs distillation.} This contradicts the conventional idea in uniform mapping. For instance, the 4-layer architecture only distills the third and fourth layers with the mapping strategy (0, 0, 5, 10)~\footnote{Recall that `0' implies that the layer does not learn from any layer from the teacher network.}, and similarly for the 6-layer architecture with two layers being distilled.
\textbf{2) Intermediate layers of teacher networks are more preferred for distillation. }
We conjecture that the top layers tend to be more relevant to the pre-training tasks~(e.g., masked language modeling) and can not be easily transferred to downstream tasks. Coincidentally, this observation is consistent with the conclusion that the middle layers of BERT are overall the most transferable~\cite{liu-etal-2019-linguistic}. 

\subsubsection{Comparisons with SOTA Methods} We compare our distilled model with ELM to state-of-the-art methods, including directly Fine-tuning~\cite{tsai2019small}, DistilBERT~\cite{sanh2019distilbert},  task-agnostically distilled TinyBERT~\cite{jiao2019tinybert} dubbed as Ta-TinyBERT and MiniLM~\cite{wang2020minilm}. 

In Table~\ref{tab:sota_compare_dev}, we can find that our models achieve the new state-of-the-art results on GLUE tasks given limited model size. For instance, under the 6-layer model architecture setting, we achieve 54.2 mcc on CoLA, which is a clear margin of gain over the rest approaches. The average performance is improved to 82.9, only 0.5 lower than the BERT$_{\rm BASE}$ teacher model. Moreover, our model performs consistently better than the baselines with the same 4-layer model architecture on all the GLUE tasks and obtains an improvement of at least 4.0\% on average. In addition, comparing to Fine-tuning that pre-training a model with the masked language model task, our approach performs significantly better on all the reported tasks, demonstrating the role of knowledge distillation. In comparison with Ta-TinyBERT, the improvement is largely attributed to the layer mapping strategy found by ELM since all the settings are consistent, which demonstrates that a different mapping strategy can indeed make a difference.

We also submitted our model predictions to the official GLUE evaluation server to obtain results on the test set~\footnote{\url{https://gluebenchmark.com/}}, as shown in Table~\ref{tab:sota_compare_test}. The results demonstrate the supremacy of our model over 1) directly pre-training a compact model, and 2) other KD-based compression approaches.


\begin{table}[]
	\caption{The evolutional process of ELM. The best \textit{gene} in each generation is shown in binary codes. Max, Min, Avg and Std indicate the highest, lowest, average and standard deviation of fitness scores in the corresponding generations.}
	\label{table: proxy experiment details}
	\centering
	\scalebox{0.75}{
		\begin{tabular}{c|cccc|l}
			\toprule
			Gen &  Max  &  Min  &  Avg  & Std & Best Gene   \\ \midrule
			\multicolumn{6}{l}{\it{Student model with 4 layers}}                         \\ \midrule
			\it{\#1}  & 81.41 & 78.91 & 80.35 & 0.62  & 000-000-011-101         \\ \midrule
			\it{\#2}  & 81.51 & 79.65 & 80.70 & 0.56  & 000-000-100-101         \\ \midrule
			\it{\#3}  & 81.51 & 80.20 & 81.10 & 0.42  & 000-000-100-101         \\ \midrule
			\it{\#4}  & 82.02 & 80.67 & 81.25 & 0.32  & 000-000-010-101         \\ \midrule
			\it{\#5}  & 82.02 & 80.51 & 81.27 & 0.31  & 000-000-010-101         \\ \midrule
			\multicolumn{6}{l}{\it{Student model with 6 layers}}                         \\ \midrule
			\it{\#1}  & 84.71 & 83.14 & 83.77 & 0.40  & 000-000-000-011-000-101 \\ \midrule
			\it{\#2}  & 84.71 & 83.46 & 84.04 & 0.34  & 000-000-000-011-000-101 \\ \midrule
			\it{\#3}  & 84.98 & 83.48 & 84.35 & 0.40  & 000-000-011-000-000-101 \\ \midrule
			\it{\#4}  & 85.30 & 84.00 & 84.58 & 0.35  & 000-100-000-000-000-101 \\ \midrule
			\it{\#5}  & 85.30 & 84.06 & 84.68 & 0.33  & 000-100-000-000-000-101 \\ \midrule
	\end{tabular}}
\end{table}

\subsection{Further Discussions}
In this section, we provide further analysis of the proposed ELM framework. We take a close look at the evolution process and verify whether the proxy setting is sufficient to perform a valid search given the advantage of efficiency. Additionally, we study the convergence behavior of ELM as well as its generalization to task-specific distillation.


\label{subsection:effectiveness_proxy_task}




\paragraph{\bf Evolution Process of ELM}

\begin{figure}
	\centering
	\includegraphics[scale=0.19]{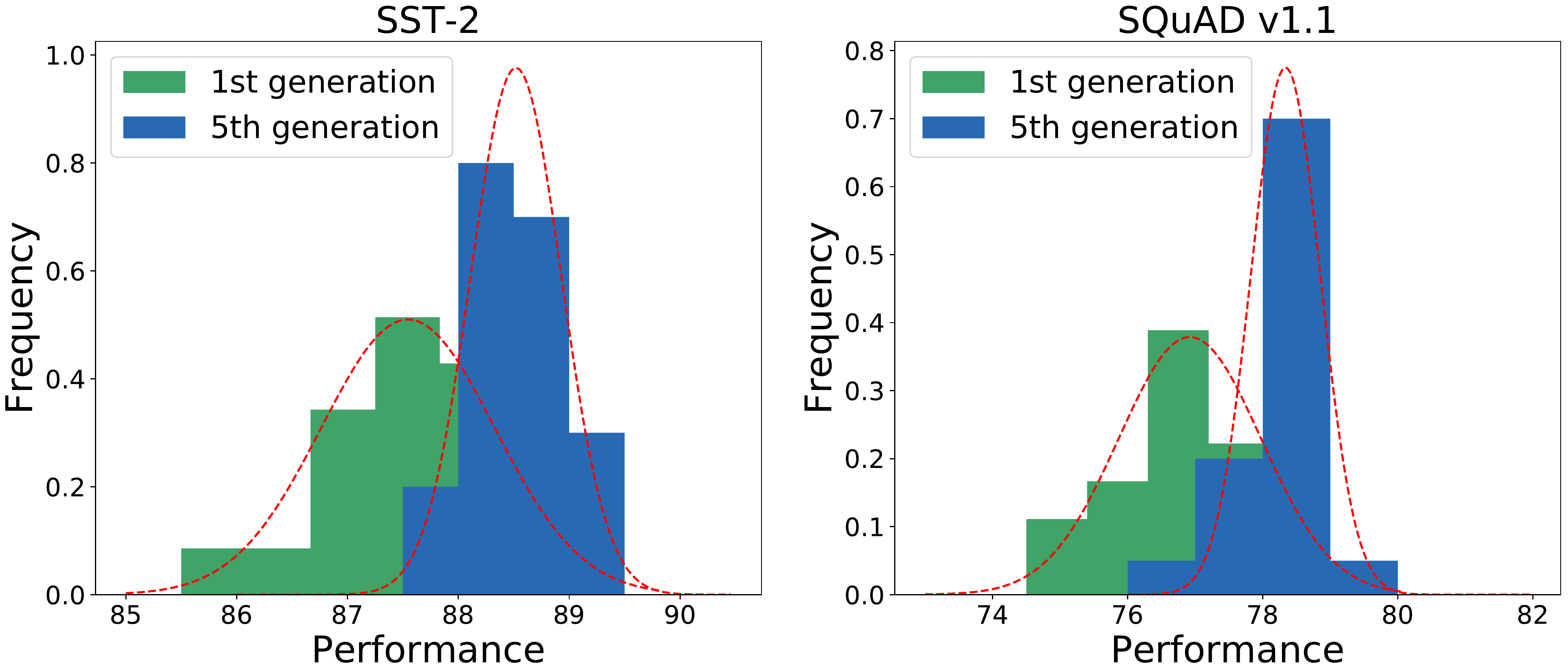}
	\caption{Performances of distilled students~(4 layer with 312 hidden size) in different generations on the datasets of SST-2 and SQuAD v1.1 under the proxy setting}
	\label{figure:evolve_perf}
\end{figure}

The statistics of the evolved layer mappings are shown in Table~\ref{table: proxy experiment details}. We can observe that the performance of task-agnostic BERT distillation highly depends on the factor of layer mapping. In each generation, there is a noticeable performance gap between the best and worst layer mapping. In the first randomly initialized generation, where the best mapping outperforms the worst one by 2.5\% on average for the 4-layer student. Moreover, after the evolution of each generation, the average performance of the current generation consistently outperforms their parent generation.
We also plot the performance distribution of the 1-st and 5-th generations in Figure~\ref{figure:evolve_perf}, and it can be found an improvement shift of performance. These results confirm that our ELM continues to improve in each evolution and finally can converge to better layer mappings strategies.

\begin{figure*}
	\centering
	\subfigure[SST-2]{\includegraphics[width=0.325\textwidth]{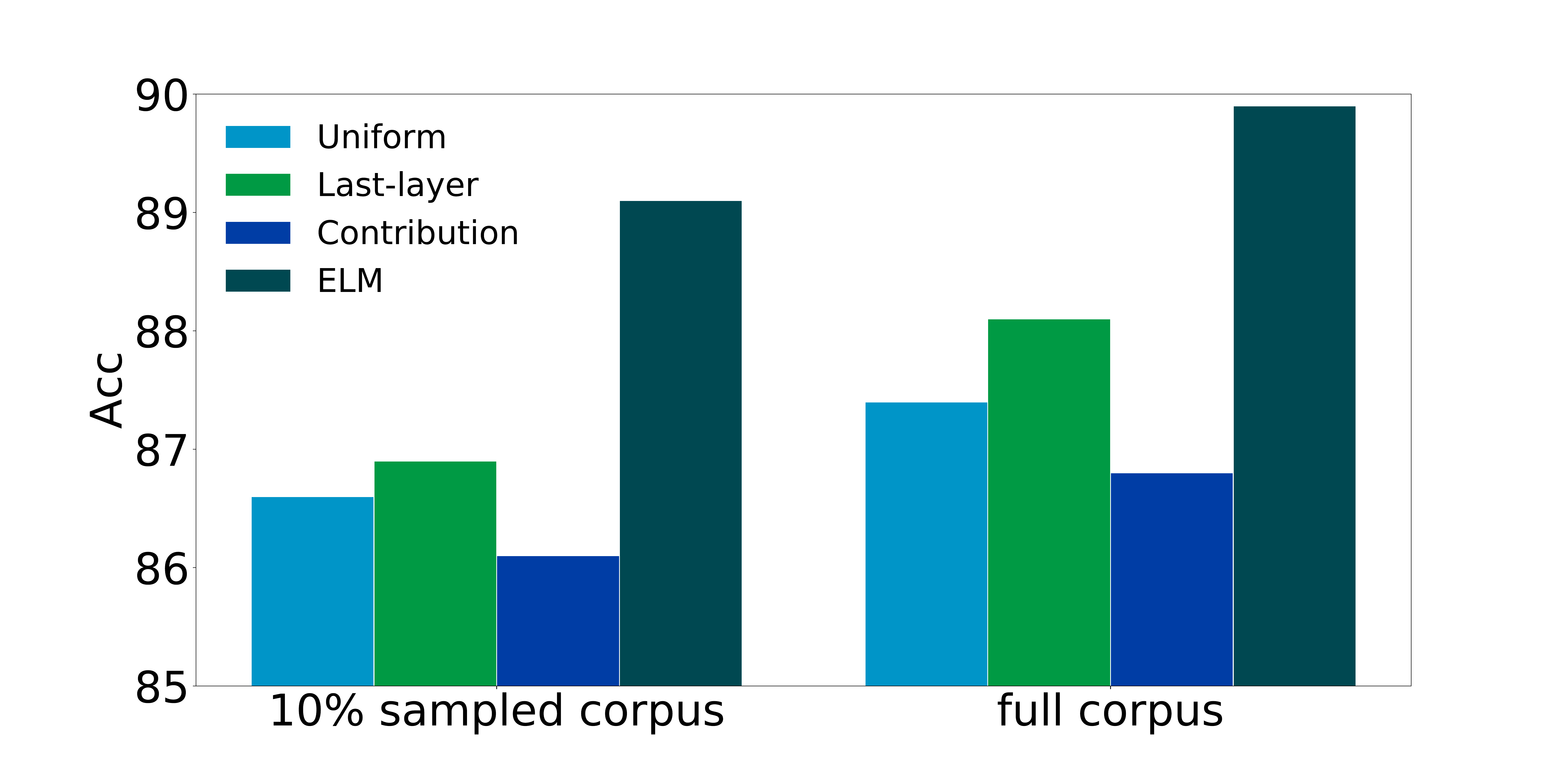}} 
	\subfigure[MNLI]{\includegraphics[width=0.325\textwidth]{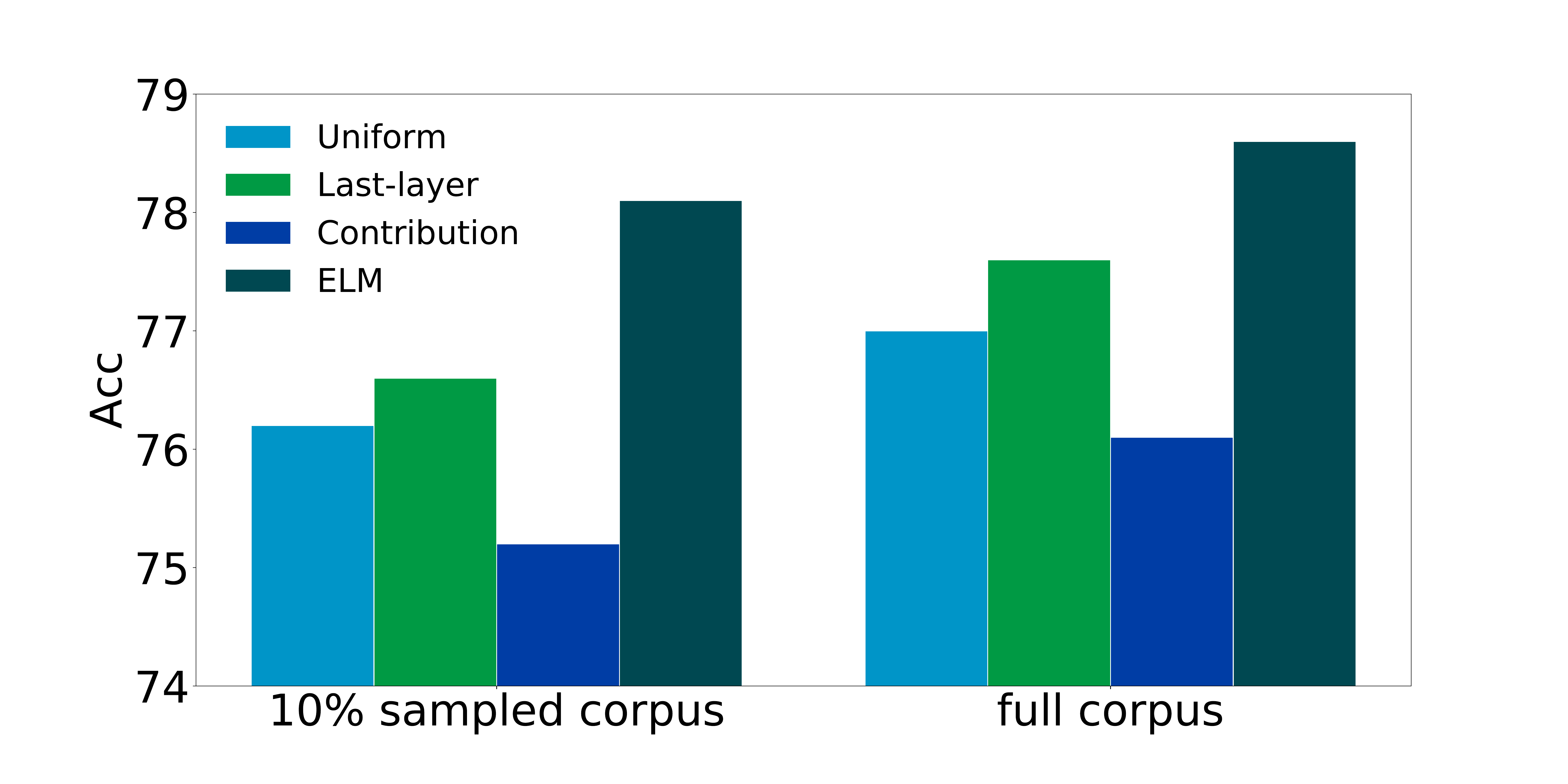}}
	\subfigure[SQuAD v1.1]{\includegraphics[width=0.325\textwidth]{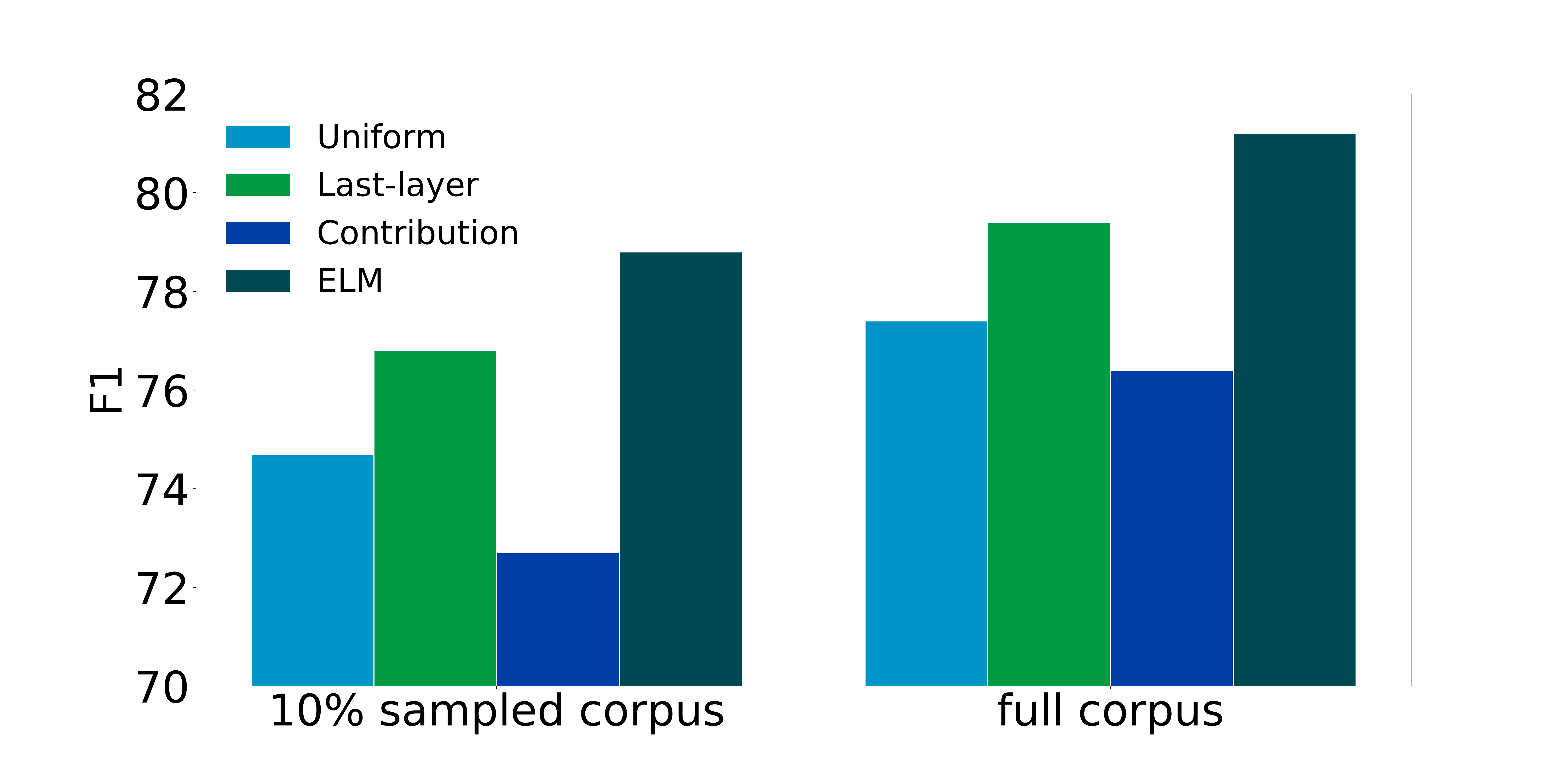}}
	\caption{The performance of distilled students~(4 layer with 312 hidden size) with different layer mappings under the proxy setting (only 10\% sampled training corpus are used) and the normal training setting with all the training corpus.}
	\label{fig:relative_performance_order}
\end{figure*}

\paragraph{\bf Effectiveness of the Proxy GA} 
Recall that our ELM is performed on a proxy setting where only 10\% sampled training corpus are used. We now turn to investigate whether such proxy setting is representative to reflect the true differences among mapping strategies.
In Figure~\ref{fig:relative_performance_order}, we show the results under the proxy setting and the normal training process with the full corpus. We can see that the Accuracy~(Acc) or F1 of both settings have the {\it same relative performance order}. Therefore, it is sufficient to rely on the proxy setting to decide the goodness of layer mappings strategies, while significantly reducing the search cost at the same time.

\begin{figure}[t]
	\centering
	\subfigure[SST-2]{\includegraphics[width=0.235\textwidth]{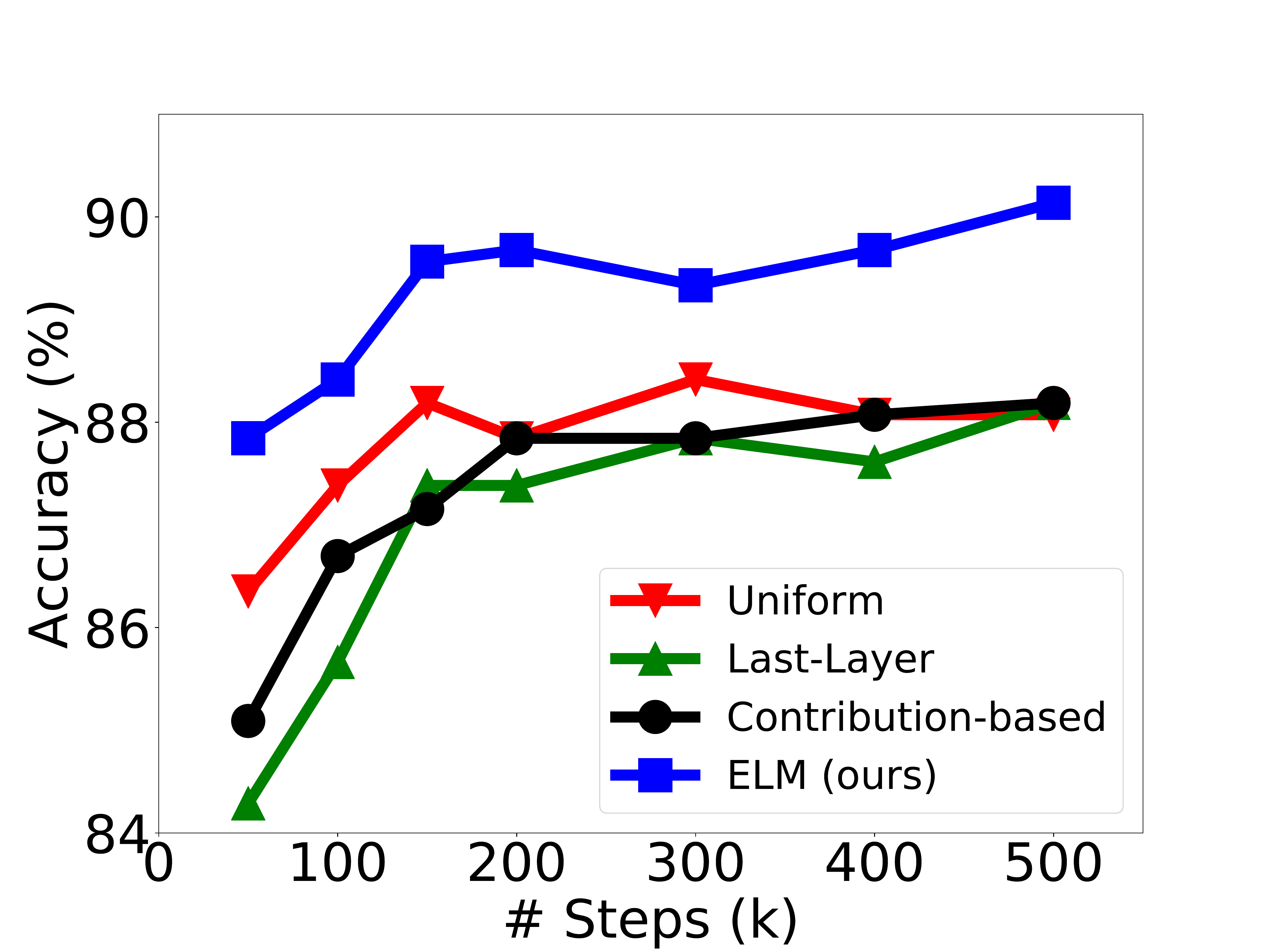}} 
	\subfigure[MNLI]{\includegraphics[width=0.235\textwidth]{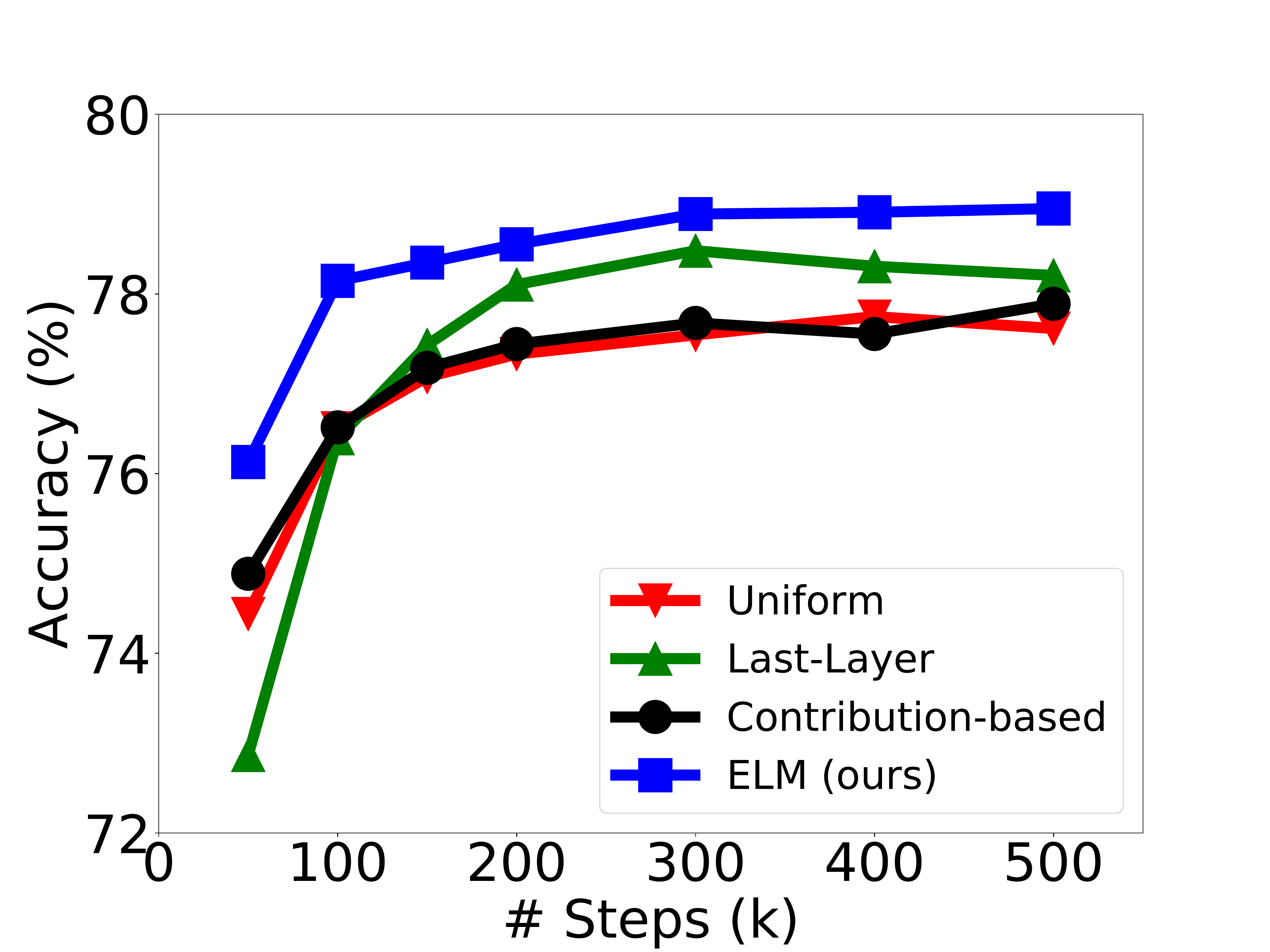}}
	\subfigure[SST-2]{\includegraphics[width=0.235\textwidth]{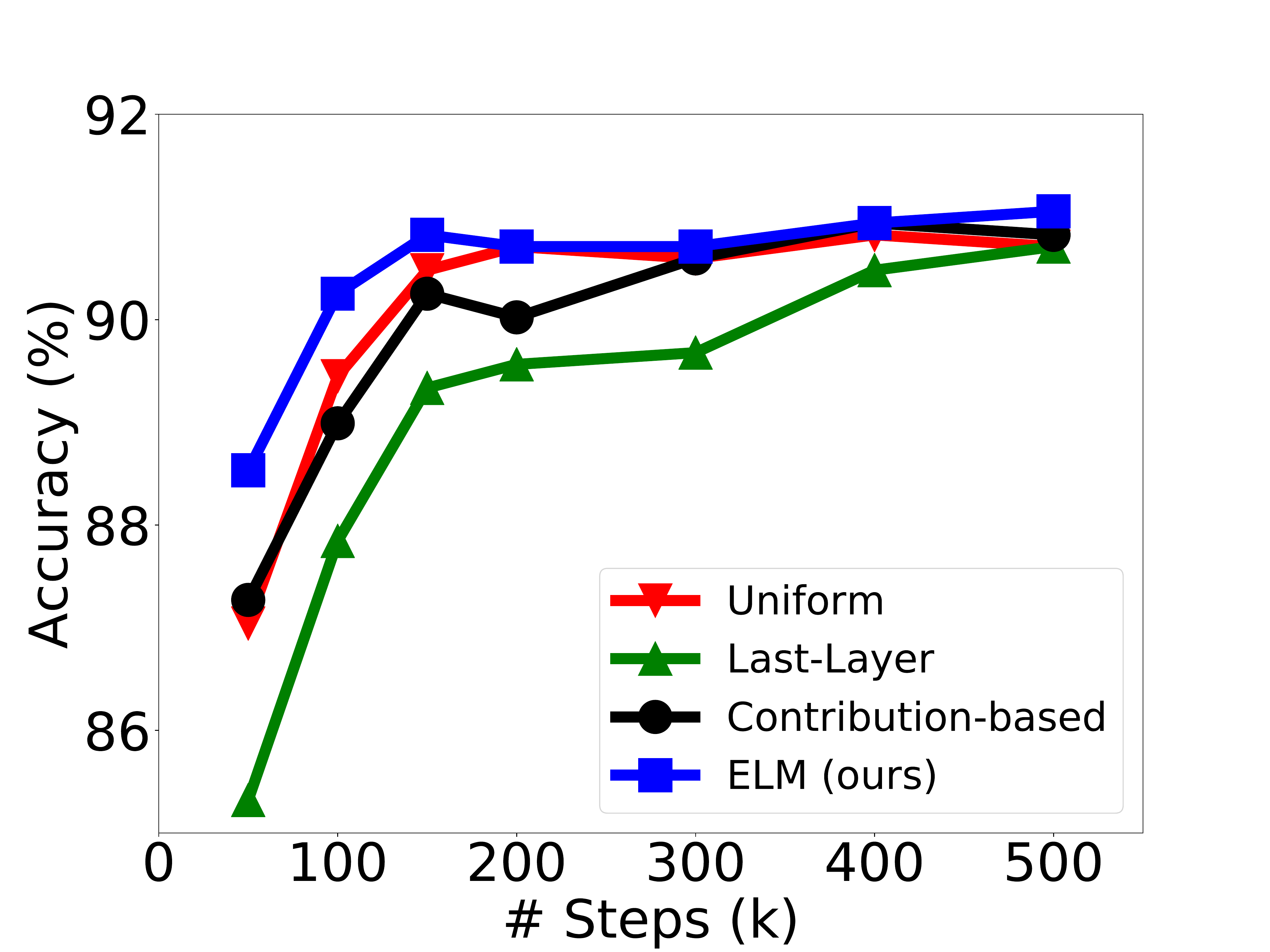}}
	\subfigure[MNLI]{\includegraphics[width=0.235\textwidth]{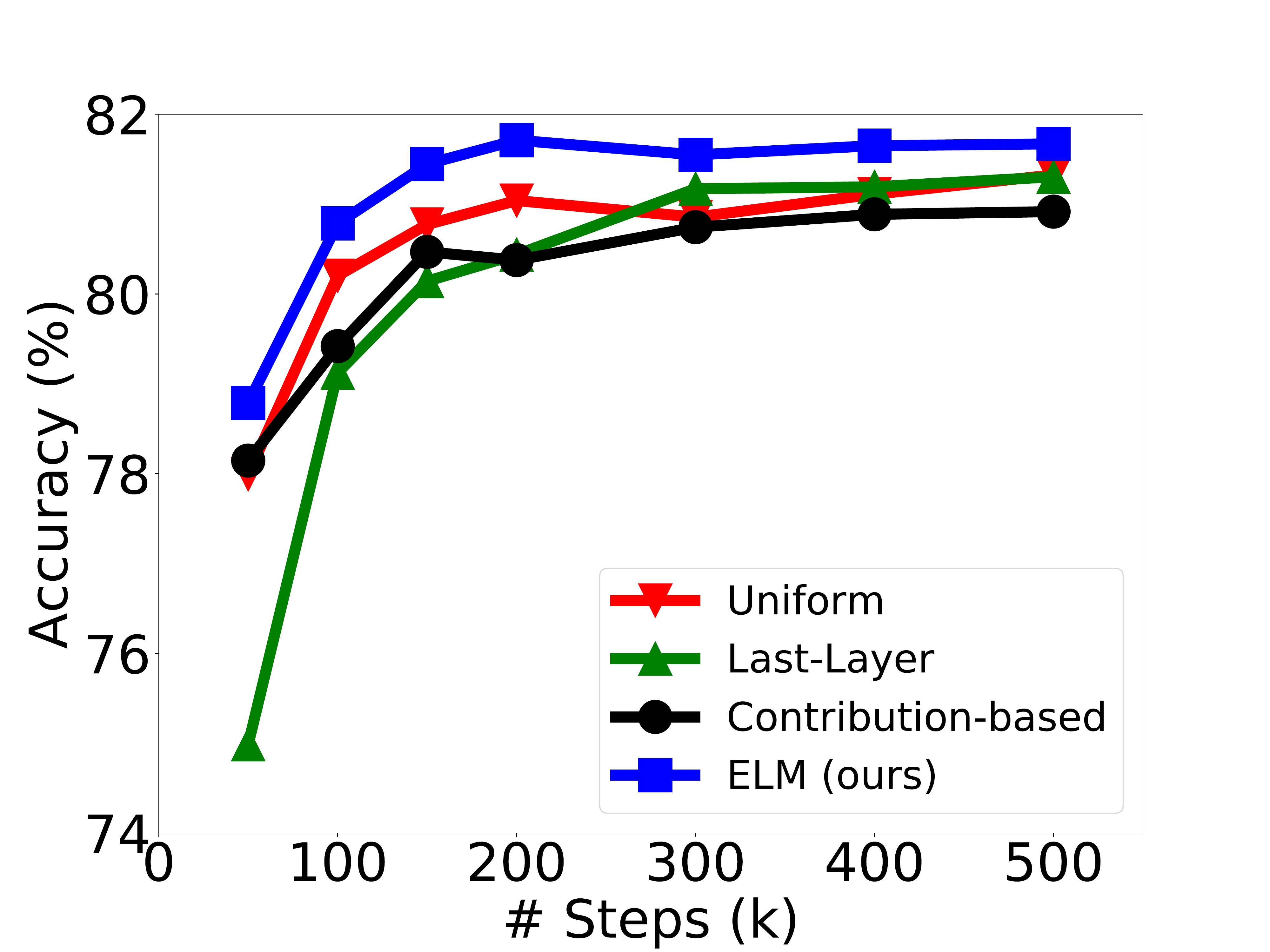}}
	\caption{Accuracies on the dev set of SST-2 and MNLI with different mapping strategies. The subfigures (a) and (b) are the results of 4-layer student models, (c) and (d) are the results of 6-layer student models.}
	\label{fig:convergence}
\end{figure}

\paragraph{\bf Convergence Curvature} We empirically study the convergence curvatures of different mapping strategies. The training curves of the 4 and 6 layer student models are shown in Figure~\ref{fig:convergence}. We can see that the ELM is consistently better than the other layer mapping strategies in the whole training stage. Our ELM can achieve comparable results with the other baselines after relatively few optimization steps~(i.e., about 150k steps), which induces that our method has the advantage of accelerating the convergence of time-consuming distillation. 
Among the three baselines, the last-layer strategy performs worst in both the SST-2 and MNLI datasets at the beginning (50k steps) but converges to similar accuracies with other baselines at the end. 

\begin{table}
	\caption{Results~(dev) of combining the proposed method with TinyBERT by replacing the uniform strategy with the optimal layer mapping. TinyBERT has an architecture of ($M$=4, $d$=312, $d_i$=1200, $h$=12).}
	\label{tab:combine_tinybert}
	\centering
	\scalebox{0.9}{
		\begin{tabular}{l|ccc}
			\toprule
			\multirow{2}{*}{Models} & SST-2 & MNLI  & SQuAD v1.1 \\
			                        & (Acc) & (Acc) &  (EM/F1)   \\ \midrule
			BERT$_{\rm BASE}$       & 93.1  & 84.2  & 80.7/88.4  \\ \midrule
			TinyBERT                & 92.7  & 82.8  & 72.7/82.1  \\ 
			TinyBERT+ELM          & 92.8  & 82.9  & 74.9/83.9  \\ 
			\bottomrule
		\end{tabular}}
\end{table}

\paragraph{\bf Improving Task-specific KD} 
We now turn to investigate whether ELMs can be extended to improve the task-specific BERT distillation. 
It is known that after completing task-agnostic distillation, it can further boost the performances to perform task-specific KD on downstream tasks~\cite{jiao2019tinybert}. 
We replace the uniform layer mapping of TinyBERT to the optimal one, found by ELM, and denote it as TinyBERT+ELM. Specifically, we use the 4-layer student model as a study case and conduct experiments on three downstream tasks. The detailed results are shown in Table~\ref{tab:combine_tinybert}. It is intriguing to notice that the EM value of TinyBERT on the challenging SQuAD v1.1 task can be further improved by 2.2\%, by directly using our optimal layer mapping. This indicates that even though the layer mapping is searched for task-agnostic distillation, it can be directly transferred to task-specific distillation to further improve the SOTA distillation methods.

\section{Conclusion}
In this paper, we proposed a layer mapping search framework, ELM, to improve the task-agnostic BERT distillation. Extensive experiments on the evaluation benchmarks show that the strategy of layer mapping has a significant effect on the distillation and the optimal layer mapping found by ELM consistently outperforms widely-used ones. In future work, we will apply the proposed search method to more scenarios related to BERT distillation, such as searching for better objectives for BERT distillation and better architectures for student models under a set of limited resources

\bibliography{emnlp2020}
\bibliographystyle{acl_natbib}

\appendix
\section*{Appendix}

\section{Chinese Experiments}
\label{sec:appx}
We further explore whether our GA based layer search algorithm is applicable to Chinese tasks.

\subsection{Datasets and Setup}
Similar to the settings of English tasks, we perform the task-agnostic BERT distillation on the Chinese Wikipedia corpus~(1.3G) and evaluate different layer mappings on the ChineseGLUE\footnote{\url{https://github.com/ChineseGLUE/ChineseGLUE}} tasks. In this work, we select three typical tasks, including one sentence entailment classification task CMNLI, one keyword recognition task CLS, and one reading comprehension task CMRC 2018~\cite{cui-etal-2019-span}.

We use the strategies of Uniform and Last-Layer as the baselines. The pre-trained Chinese BERT$_{\rm BASE}$ is used as the teacher, and a model of architecture ($M$=4, $d$=312, $d_i$=1200, $h$=12) is used as the student network. Compared to the English training corpus, the Chinese Wikipedia is relatively small, we perform distillation with a maximum sequence length of 128 for nine epochs, and 512 for one more epoch. We use the evaluation scripts provided by ChineseGLUE for fine-tuning, and keep the default hyper-parameters during the whole searching process.

\subsection{Results and Analysis}

\begin{table}[]
	\caption{Results of 4-layer student ($M$=4, $d$=312, $d_i$=1200, $h$=12) with different layer mappings in Chinese tasks.}
	\label{tab:chinese_main_results}
	\centering
	\scalebox{0.7}{
		\begin{tabular}{l|c|ccc|c}
			\toprule
			Strategy            & Layer Mapping &     CMNLI     &      CSL      &   CMRC 2018   &      Avg      \\ \midrule
			Uniform             &  (3,6,9,12)   &     72.7      &     72.1      &     48.2      &     64.3      \\
			Last-Layer          &  (0,0,0,12)   &     73.0      &     71.7      &     51.7      &     65.5      \\
			\textbf{ELM (ours)} &  (1,4,6,11)   & \textbf{73.5} & \textbf{73.9} & \textbf{55.6} & \textbf{67.7} \\ \bottomrule
	\end{tabular}}
\end{table}

The results of the Chinese experiment are presented in Table~\ref{tab:chinese_main_results}. We have a similar observation as in English experiments. The strategy of layer mapping still dramatically affects the performances of BERT distillation. The proposed ELM outperforms the widely used baselines by at least 2.2\% on average, especially on the challenging reading comprehension CMRC task. The optimal layer mapping for Chinese tasks is (1, 4, 6, 11), which does not strictly match that, i.e. (0, 0, 5, 10), of English tasks. However, we can observe that they still possess some similar properties, e.g., both the two optimal layer mappings contain the more transferable middle layers~(layer `6' and `5'), the last layer of the teacher network is not selected. Three layers of the student network will learn from a layer of the bottom 6 layers of the teacher network. It is natural to ask the question of whether there exists an optimal ELM for multilingual tasks, which requires more computational resources, and we leave it as future work. 

\end{document}